# Deep Predictive Video Compression with Bi-directional Prediction

Woonsung Park    Munchurl Kim

School of Electrical Engineering
Korea Advanced Institute of Science and Technology (KAIST), Daejeon, Korea

{pys5309, mkimee}@kaist.ac.kr

**Abstract**

*Recently, deep image compression has shown a big progress in terms of coding efficiency and image quality improvement. However, relatively less attention has been put on video compression using deep learning networks. In the paper, we first propose a deep learning based bi-predictive coding network, called BP-DVC Net, for video compression. Learned from the lesson of the conventional video coding, a B-frame coding structure is incorporated in our BP-DVC Net. While the bi-predictive coding in the conventional video codecs requires to transmit to decoder sides the motion vectors for block motion and the residues from prediction, our BP-DVC Net incorporates optical flow estimation networks in both encoder and decoder sides so as not to transmit the motion information to the decoder sides for coding efficiency improvement. Also, a bi-prediction network in the BP-DVC Net is proposed and used to precisely predict the current frame and to yield the resulting residues as small as possible. Furthermore, our BP-DVC Net allows for the compressive feature maps to be entropy-coded using the temporal context among the feature maps of adjacent frames. The BP-DVC Net has an end-to-end video compression architecture with newly designed flow and prediction losses. Experimental results show that the compression performance of our proposed method is comparable to those of H.264, HEVC in terms of PSNR and MS-SSIM.*

1. Introduction

Conventional video codecs such as AVC/H.264 [1], HEVC [2] and VP9 [3] have shown significantly improved coding efficiencies, especially by enhancing their temporal prediction accuracies for the current frame to be encoded using its adjacent frames. In particular, there are three types of frames used in video compression: I-frame (intra-coded frame) that is compressed independently of its adjacent frames; P-frame (predicted frame) that is compressed through the forward prediction using motion information; and B-frame (bidirectional predicted frame) that is compressed with bidirectional prediction for the current frame. In perceptive of coding efficiency, B-frame coding provides the highest coding efficiency compared to the I-frame and P-frame coding methods.

Deep learning based approaches have recently shown significant performance improvement in image processing. Especially, in the field of low-level computer vision, intensive research has been made for deep learning based image super-resolution [4-7] and frame interpolation [8-12]. In addition, there are many recent studies on image compression using deep learning [13-22] which often incorporate auto-encoder based end-to-end image compression architectures by attempting to improve compression performance. These works reported outperformed results of coding efficiency compared to the traditional image compression methods such as JPEG [23], JPEG2000 [24], and BPG [25]. While the image compression tries to reduce only spatial redundancy around the neighboring pixels with limited coding efficiency, traditional video compression can achieve significant compression performance because it can take advantage of temporal redundancy among neighboring frames. Also, by exploiting the temporal redundancy, deep learning based video compression has been studied in two main directions: First, some components (or coding tools) in the conventional video codecs are replaced with deep neural networks. For example, Park and Kim [26] first tried to improve compression performance by replacing the in-loop filters of HEVC with a CNN-based in-loop filer. Cui *et al* [27] proposed intra-prediction method with CNN in HEVC to improve compression performance. Zhao *et al* [28] replaced the bi-prediction strategy in HEVC with CNN to improve coding efficiency; Second, there are studies to improve the compression performance by using auto-encoder based end-to-end neural network architectures as a completely different video coding paradigm [29-31].

Although deep learning based image compression has been intensively studied, deep learning based video compression has drawn less attention. In this paper, we first propose an end-to-end deep predictive video compression scheme with optical flow-based bi-predictive prediction, called BP-DVC Net. The contribution of our proposed

BP-DVC Net has the following main contributions:

i) We first incorporate a bi-directional prediction network into the BP-DVC Net using optical flow information from future and past frames to the current frame to be encoded. The resulting residues from the bidirectional prediction are efficiently encoded.

ii) The BP-DVC Net incorporates forward and backward optical flow estimation networks in both encoder and decoder sides to predict the current frame with the resulting residual information as small as possible. Furthermore, the bi-directional optical flows are not transmitted to the decoder sides so that high efficiency can be achieved. For predictive video coding, precise optical flow estimation must be possible between two frames in a far-away distance with each group of pictures (GOP), which often fails when using the previous deep learning based naïve optical flow estimation networks. In order to obtain accurate optical flow in such cases, an optical flow refinement network is proposed based on a U-Net structure with a novel optical flow loss. Therefore, our proposed optical flow estimation network can yield accurate optical flow fields that helps increase the coding efficiency of deep predictive video coding.

iii) We propose an entropy coding network that exploits the temporal context information among the feature maps of the neighboring frames to obtain improved coding efficiency.

This paper is organized as follows: Section 2 introduces the related works with deep neural network-based image and video compression, optical flow estimation and frame interpolation; In Section 3, we introduce our proposed deep end-to-end video compression network based on bi-directional prediction, which is called BP-DVC Net; Section 4 presents the experimental results to show the effectiveness of our proposed BP-DVC Net compared to the conventional video compression codecs; Finally, we conclude our work in Section 5.

2. Related Work

Both conventional image compression (such as JPEG, JPEG2000, and BPG) and video compression (AVC/H.264, HEVC, and VP9) have shown high compression performance by exploiting spatial redundancy information and temporal redundancy information, respectively. Recently, deep learning-based image compression and video compression methods have been actively studied. The key element that brings up high coding efficiency in video coding is temporal prediction to reduce temporal redundancy. Therefore, we review deep learning-based optical flow networks or frame interpolation networks that can be used for such a prediction purpose.

2.1. Deep learning-based image compression

Unlike conventional image compression based on transform coding, recent deep learning-based image compression methods often adopt auto-encoder structures that perform nonlinear transforms. First, there are several works on image compression using Long Short Term Memory (LSTM)-based auto-encoders [13-15] where a progressive coding concept is used to encode the difference between the original image and the reconstructed image in the LSTM-based auto-encoder structure in a progressive manner. In addition, there are studies on image compression using convolutional neural network (CNN) based auto-encoder structures by modeling the feature maps of the bottleneck layers for entropy coding [16-22]. Ballé *et al.* [16] showed good compression performance by applying a nonlinear activation function, called generalized divisive normalization (GDN) with non-parametric models. Thesis *et al.* [22] have improved the performance of entropy coding through the assumption that the feature map of the bottleneck layer is based on a Gaussian scale mixture model with six zero-mean Gaussian models. Their models in [16, 22] outperform the conventional image codecs such as JPEG2000. The both methods in [16, 22] have not taken into account the input-adaptive entropy models. Ballé *et al.* [17] introduced an input-adaptive entropy model that estimates the scales of the representations depending on the input. Lee *et al.* [19] have proposed a context-adaptive entropy model for image compression which uses two types of contexts: bit-consuming context and bit-free context. Their models in [17, 19] outperformed the conventional image codecs such as BPG. In our BP-DVC Net, we adopt an LSTM-based auto-encoder structure used in [15] as the baseline structure with progressive coding.

2.2. Deep learning based video compression

Similar to the deep learning-based image compression, there are two main types of video compression research using deep learning: The first is to replace the existing components of the conventional video compression codecs with deep neural networks (DNN), which is not based on end-to-end learning schemes. For example, there are some works to replace in-loop filters with deep neural networks [26, 32-34], and post-processing to enhance the resulting frames of the conventional video codecs [35, 36]. The intra/inter predictive coding modules have also been substituted with DNN modules for video coding [27, 28]; And, the second is an auto-encoder based video compression architecture that constructs the entire structure in terms of deep neural networks without the coding tools of conventional video codecs involved. The CNN-based end-to-end auto-encoder networks have been proposed for P-frame prediction which only use a previous frame to predict the current frame [30, 31]. Xu *et al.* proposed an LSTM auto-encoder based video compression method to improve the coding efficiency [29].

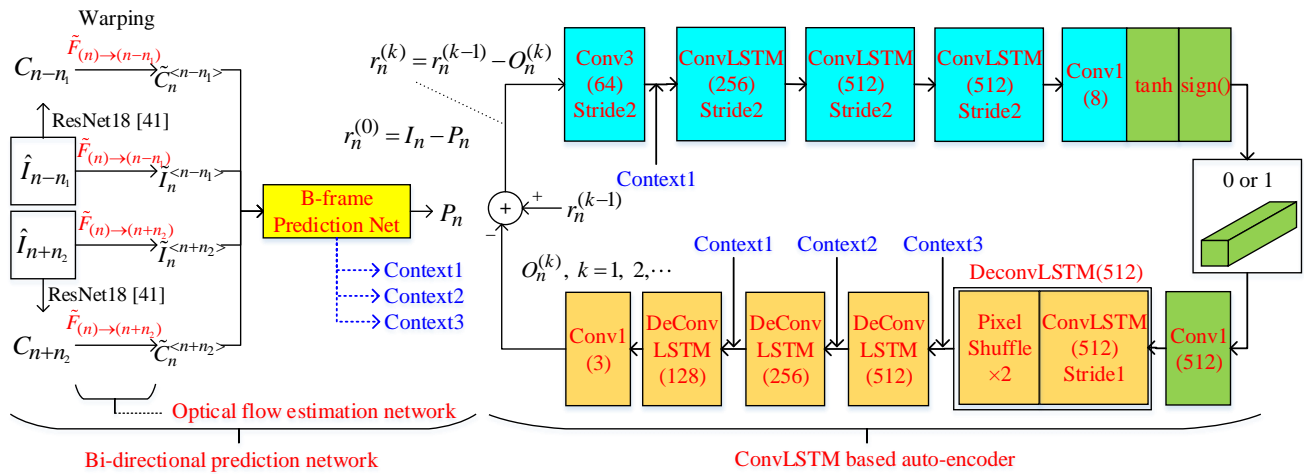

Figure 1. Overall Architecture of the proposed deep predictive video codec with a bidirectional prediction network (BP-DVC Net). The boxes with red-colored numbers and names indicate specific network modules with the convolution filter parameters that cab be trained in end-to-end manners; The context information in blue color is fed into some network modules as concatenated input into both the encoder and decoder of the auto-encoder).

### 2.3. Deep learning based optical flow estimation

Optical flow can be used for predictive coding in video codecs and frame interpolation for frame rate up-conversion. There have been many studies related to optical flow estimation using deep neural networks. Dosovitskiy *et al*. [37] proposed the CNN based models for optical flow estimation, called FlowNetS and FlowNetC that are based on U-Net structures [38]. Ranjan *et al*. [39] introduced SpyNet that uses a spatial pyramid network and warps the second image to the first image with the initial optical flow. Also, the PWC-Net [40] was introduced with a learnable feature pyramid structure that uses the estimated current optical flow to warp the CNN feature maps of the second image. Then, the loss between the warped feature maps and the feature maps of the first image is used to improve the accuracy of optical flow by CNN. Their model outperformed all previous optical flow methods. Since they use the feature pyramid structures, the optical flow estimation is robust to large motion over other deep neural network based optical flow methods. We also use PWC-Net as a pre-trained optical flow estimation network and its output is used as the initial optical flow for bidirectional prediction.

### 2.4. Deep learning based frame interpolation

Recent DNN-based frame interpolation methods include convolution filtering-based frame interpolation [8, 9], phase-based frame interpolation [10], and optical flow-based interpolation [11, 12]. The convolution filtering-based frame interpolation predicts frames between adjacent frames through convolution filtering operation without using optical flow. The phase-based frame interpolation uses a DNN to reduce the reconstruction loss in the phase domain rather than in the image domain. Finally, the optical flow-based interpolation generates the frames between two frames through a DNN after warping with optical flow between two frames. Especially, it is important to obtain optical flow as accurate as possible because the accuracy of estimated optical flow has a great influence on the performance of the frame interpolation in the optical flow based-frame interpolation. In this paper, we propose an optical flow-based bidirectional prediction network and add a DNN-based optical flow refinement subnet to obtain better optical flow with enhanced accuracy.

## 3. Proposed BP-DVC Net

### 3.1. Overall Architecture

Fig. 1 shows the overall architecture of our BP-DVC Net. As shown in Fig. 1, the BP-DVC Net consists of an optical flow estimation network based on one U-Net, a bidirectional prediction network based on another U-Net, and a Conv-LSTM based auto-encoder network, all of which can be trained in an end-to-end manner. In order to compress the current frame $I_n$, $I_n$ is first predicted from two previously encoded neighboring frames $\hat{I}_{n-n_1}$ and $\hat{I}_{n+n_2}$ using the bi-directional prediction network. The selection of the neighboring frames $\hat{I}_{n-n_1}$ and $\hat{I}_{n+n_2}$ is determined depending on the temporal hierarchical levels in the hierarchical B-picture prediction structure. Once the prediction $P_n$ of $I_n$ is calculated, the residue, $r_n^{(0)} = I_n - P_n$ is fed into the convolutional LSTM (ConvLSTM)-based auto-encoder network as the input at iteration 0. Then the ConvLSTM-based auto-encoder network is trained to output $O_n^{(k)}$ toward progressively reducing the residue $r_n^{(k)} = r_n^{(k-1)} - O_n^{(k)}$. The details of the bidirectional prediction network and the optical flow estimation network and the ConvLSTM-based auto-encoder network are described in Section 3.3, 3.4 and 3.5, respectively.

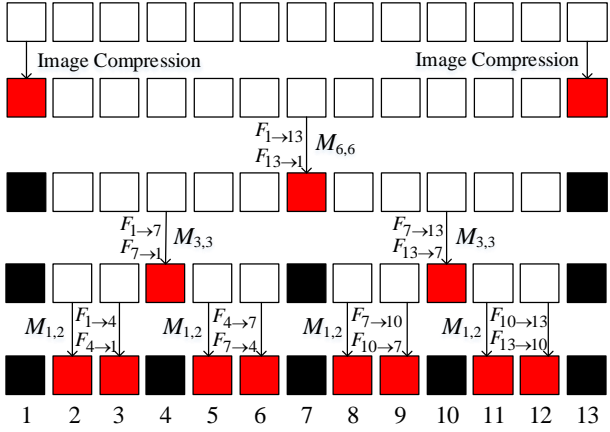

Figure 2. Proposed hierarchical B-frame structure for bi-predictive optical flow estimation and bi-directional frame prediction in the BP-DVC Net.

### 3.2. Proposed hierarchical B-frame structure for bi-directional predictive coding with bi-predictive optical flow estimation

Fig. 2 shows a hierarchical B-frame structure for bi-predictive optical flow estimation and bi-directional frame prediction in the BP-DVC Net. In Fig. 2, the hierarchical B-frame structure has a group of pictures (GOP) size with 12 frames. In our optical flow estimation network, a hierarchical bi-predictive optical flow estimation scheme is used in a GOP. In Fig. 2, *frames* 1 and 13 are first encoded. Then, *frame* 7 is bi-predicted using the forward and backward optical flow estimations ($F_{1\rightarrow13}$ and $F_{13\rightarrow1}$) from *frame* 1 to *frame* 13 and vice versa. It should be noted that the previous hierarchical bi-predictive optical flow estimation [29] was made from *frame* 1 to *frame* 7 ($F_{1\rightarrow7}$) and from *frame* 13 to *frame* 7 ($F_{13\rightarrow7}$). In this case, both the optical flow fields must be transmitted to the decoder sides. However, our hierarchical bi-predictive optical flow estimation scheme does not require to transmit the estimated optical flows to decoder sides. In Fig. 2, $M_{n_1,n_2}$ indicates a bi-predictive video coding model to encode the *n*-th frame between its previously encoded ($n$-$n_1$)-th frame and its previously encoded ($n$+$n_2$)-th frame where $n > n_1$. For simplicity, we assume $M_{n_1,n_2} = M_{n_2,n_1}$ where $M_{n_1,n_2}$ takes as input both forward and backward optical flows estimated by a single optical flow estimation model that is trained bi-directionally between ($n$-$n_1$)-th frames and ($n$+$n_2$)-th frames. We use three bi-predictive video coding models such as $M_{1,2}$, $M_{3,3}$ and $M_{6,6}$ in our experiments.

### 3.3. Proposed Bi-directional prediction network

The proposed bi-directional prediction network consists of two main networks: (i) The first one is an optical flow estimation network based on a U-Net structure, yielding two refined optical flows to warp two neighboring frames to the current frame; (ii) The other one is a bi-directional prediction network based on another U-Net structure that generates a prediction of the current frame by warping the neighboring two frames to the current frame based on the refined optical flows.

#### 3.3.1 Proposed Optical flow estimation network

In order to avoid the transmission of the optical flow to decoder sides for high coding efficiency, the forward and backward optical flows from the current *n*-th frame $I_n$ to its previously encoded ($n$-$n_1$)-th frame $\hat{I}_{n-n_1}$ and to its previously encoded ($n$+$n_2$)-th frame $\hat{I}_{n+n_2}$ are not directly computed but is estimated using the forward and backward optical flows between $\hat{I}_{n-n_1}$ and $\hat{I}_{n+n_2}$. Fig. 3 shows the architecture of our proposed optical flow estimation network. It first takes as input the initial forward and backward optical flows, $F_{(n-n_1)\to(n+n_2)}$ and $F_{(n+n_2)\to(n-n_1)}$, estimated between by a pre-trained PWC-Net [40] that is robust to large motion estimation. Then, based on $F_{(n-n_1)\to(n+n_2)}$ and $F_{(n+n_2)\to(n-n_1)}$, we compute the forward and backward optical flows, $\breve{F}_{(n)\to(n-n_1)}$ and $\breve{F}_{(n)\to(n+n_2)}$, from $I_n$ to $\hat{I}_{n-n_1}$ and to $\hat{I}_{n+n_2}$ as follows:

$$\breve{F}_{(n)\to(n-n_1)} = -1/4\left[F_{(n-n_1)\to(n+n_2)} - F_{(n+n_2)\to(n-n_1)}\right]$$
$$\breve{F}_{(n)\to(n+n_2)} = 1/4\left[F_{(n-n_1)\to(n+n_2)} - F_{(n+n_2)\to(n-n_1)}\right]$$
$$\text{for } M_{3,3} \text{ and } M_{6,6} \quad (1)$$

$$\breve{F}_{(n)\to(n-n_1)} = -(2/9)F_{(n-n_1)\to(n+n_2)} + (1/9)F_{(n+n_2)\to(n-n_1)}$$
$$\breve{F}_{(n)\to(n+n_2)} = (4/9)F_{(n-n_1)\to(n+n_2)} - (2/9)F_{(n+n_2)\to(n-n_1)}$$
$$\text{for } M_{1,2} \quad (2)$$

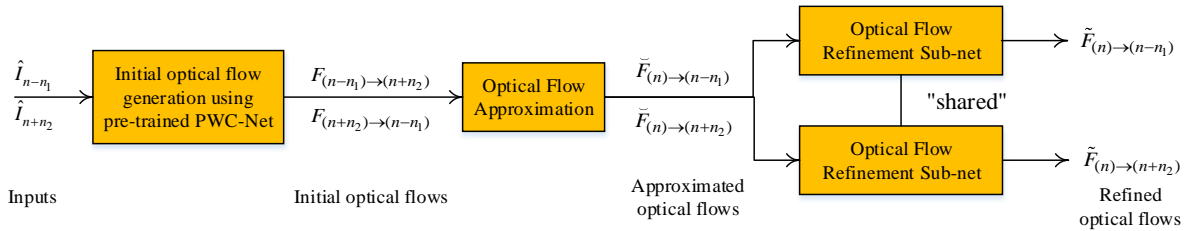

Figure 3. The proposed optical flow estimation network architecture.

It is noted that the optical flows in (1) and (2) are approximates assuming motion is smooth as in [12]. Based on (1) and (2), the decoder can also predict the optical flows, $F_{(n)\to(n-n_1)}$ and $F_{(n)\to(n+n_2)}$ only from $\hat{I}_{n-n_1}$ and $\hat{I}_{n+n_2}$, which are available in both encoder and decoder sides. However, if the distance between two frames is long or the motion is fast, the assumption with motion smoothness becomes no longer true. Therefore, we incorporate an optical flow refinement subnet to refine the approximated optical flows in (1) and (2), which has a small-sized U-Net structure which is described in the supplemental material. The optical flow refinement subnet is trained using our proposed flow loss $l_f$ as follows:

$$l_f = \frac{1}{2N}\sum_{k=1}^{N}\sum_{i,j}\left[\begin{array}{l}\left|F_{(n)\to(n-n_1)}^k(i,j) - \tilde{F}_{(n)\to(n-n_1)}^k(i,j)\right|_1 \\ + \left|F_{(n)\to(n+n_2)}^k(i,j) - \tilde{F}_{(n)\to(n+n_2)}^k(i,j)\right|_1\end{array}\right]$$
(3)

where $\tilde{F}_{(n)\to(n-n_1)}^k$ and $\tilde{F}_{(n)\to(n+n_2)}^k$ are the outputs of the optical flow refinement subnet for the $k$-th training input patches $\breve{F}_{(n)\to(n-n_1)}^k$ and $\breve{F}_{(n)\to(n+n_2)}^k$, respectively, and are compared to their respective $k$-th optical flow ground true patches, $F_{(n)\to(n-n_1)}^k$ and $F_{(n)\to(n+n_2)}^k$. It should be noted that $F_{(n)\to(n-n_1)}^k$ and $F_{(n)\to(n+n_2)}^k$ are the optical flows directly computed from $I_n$ to $\hat{I}_{n-n_1}$ and to $\hat{I}_{n+n_2}$ by using the pre-trained PWC-Net. $N$ is the total number of training optical flow patches.

### 3.3.2 Proposed Bi-directional prediction network

Based on the refined optical flows, $\tilde{F}_{(n)\to(n-n_1)}^k$ and $\tilde{F}_{(n)\to(n+n_2)}^k$, a bi-directional prediction network is proposed to predict $I_n$ to be encoded, denoted as $P_n$, which is based on a U-net structure which is described in the supplemental material. The input to the bi-directional prediction network are the generated frames, $\tilde{I}_n^{<n-n_1>}$ and $\tilde{I}_n^{<n+n_2>}$, which are obtained by warping $\hat{I}_{n-n_1}$ and $\hat{I}_{n+n_2}$ to predict $I_n$ based on $\tilde{F}_{(n)\to(n-n_1)}^k$ and $\tilde{F}_{(n)\to(n+n_2)}^k$, respectively. The bidirectional prediction network also incorporates the contexts $C_{n-n_1}$ and $C_{n+n_2}$ as in [11]. The contexts are also warped to $I_n$ in the same way as warping $\hat{I}_{n-n_1}$ and $\hat{I}_{n+n_2}$, and the warped contexts are denoted as $\tilde{C}_n^{<n-n_1>}$ and $\tilde{C}_n^{<n+n_2>}$ [42]. Then, two warped frames ($\tilde{I}_n^{<n-n_1>}$, $\tilde{I}_n^{<n+n_2>}$) and their warped contexts ($\tilde{C}_n^{<n-n_1>}$, $\tilde{C}_n^{<n+n_2>}$) are concatenated and used as inputs to the bidirectional prediction network.

The bi-directional prediction can be trained in an end-to-end manner with the optical flow estimation network and the ConvLSTM-based auto-encoder using the prediction loss $l_p$ as

$$l_p = \frac{1}{N}\sum_{k=1}^{N}\sum_{i,j}\left|I_n^k(i,j) - P_n^k(i,j)\right|_1 \quad (6)$$

where $P_n^k$ is the $k$-th training input patch, and $N$ is the total number of training image patches.

### 3.4. ConvLSTM-based auto-encoder

Our BP-DVC Net includes a ConvLSTM-based auto-encoder structure [15, 29] that is comprised of an encoder $E$, a binarizer $B$ and a decoder $D$. This LSTM based model encodes and decodes each frame progressively during $K$ iterations. At each iteration, the implementation of the auto-encoder network can be represented as:

$$O_n^{(k)} = D(B(E(r_n^{(k-1)}))),$$
$$r_n^{(k)} = r_n^{(k-1)} - O_n^{(k)}, \quad r_n^{(0)} = I_n - P_n \quad (7)$$

where $r_n^{(k)}$ is the residue at iteration $k$ between the reconstructed image $O_n^{(k)}$ and the residue $r_n^{(k-1)}$. In order to train the ConvLSTM-based auto encoder, we use an L1 loss $l_a$ between $I_n$ and $O_n^{(k)}$ at all iterations as:

$$l_a = \frac{1}{M}\frac{1}{K}\sum_{m=1}^{M}\sum_{k=1}^{K}\sum_{i,j}\left|r_n^{m,k}(i,j)\right| \quad (8)$$

where $r_n^{m,k}$ is the residue at iteration $k$ for the $m$-th training patch, $K$ is the total number of iterations in progressive residual computation by the Conv-LSTM based auto-encoder, and $M$ is the total number of training patches.

### 3.5. Proposed entropy coding model

We proposed an entropy coding model for $M_{12}$ that utilizes the temporal context of neighboring frames. Fig. 4 shows the proposed entropy coding model for the $M_{12}$ model with 11 layers of masked 3D convolutions with 128 channels. Unlike conventional image compression, video compression exploits the temporal redundancy of the current frame with its neighboring frames. Wu *et al*. [29] implemented a 3D pixel-CNN network for entropy coding that uses only the binary codes of the current frame as in [20]. However, the proposed entropy coding model for $M_{12}$ uses the temporal context of the binary codes of the current frame with respect to that of a neighboring frame. We assume that they have temporal redundancy with each other since they are encoded from the adjacent frames with the same encoder of $M_{12}$. In order to utilize the temporal context of the binary codes, a skip connection of the binary code of an adjacent frame is added to the output of the entropy coding network for the binary code of the current frame. This makes it possible to better predict the binary code of the current frame with the aid of the binary code of the adjacent frame.

### 3.6. Comparison of BP-DVC Net with the previous method [29]

In [29], they incorporated the temporal contexts (U-net feature maps) of two neighboring frames into the auto-encoder network to compress the current frame. The two previously encoded neighboring frames are directly input to the auto-encoder with the current frame, and the temporal contexts are concatenated with the interim feature maps of the auto-encoder network to predict the current frame. However, this method has a limitation in achieving high coding efficiency because (i) the auto-encoder has to perform both the compression of the current frame and the prediction from the neighboring two frames; and (ii) a single loss is used at the output of the auto-encoder, which makes it difficult the end-to-end training of the network. On the other hand, our BP-DVC Net has a separate bidirectional prediction network, yielding the residual information that is fed into the ConvLSTM-based auto-encoder. This allows the proposed bidirectional prediction network to focus on bidirectional prediction by the proposed prediction loss in (6) and the auto-encoder to concentrate on the residue compression by the reconstruction loss in (8).

In [29], in order to warp the context information of the neighboring frames, they used estimated optical flows using Farnebäck's algorithm [43] and block motions using the same algorithm as H.264 for motion estimation. Inefficiently, they used an old optical flow method and a coarse block motion estimation method, and motion information must be transmitted to decoder sides. This limits to improve the coding efficiency. On the other hand, our optical flow estimation network has an optical flow refinement subnet for better bidirectional prediction. This leads to improved overall compression performance.

Wu *et al*. [29] used a lossless 4-channel WebP [44] for block motion and a separate deep learning based image compression model for optical flow compression and transmission to a decoder. While the Wu *et al*.'s method needs to transmit motion information that significantly takes a large portion, our bi-direction optical flow estimation network is maintained in both encoder and decoder sides so motion information used in the encoder sides can be derived at the decoder sides without necessitating the optical flow transmission, which can greatly save the bitrate.

## 4. Experiments

Our BP-DVC Net is extensively tested in terms of coding efficiency and is compared with other video coding methods. In our BP-DVC Net, the inter-prediction coding models with $M_{12}$, $M_{33}$, and $M_{66}$ are focused for training and testing. For intra coding, we used the pre-trained deep neural network based image compression model in [19].

### 4.1. Experimental conditions

**Datasets.** We train the BP-DVC Net using the Kinetics video dataset [45]. However, since this dataset is composed of already compressed video data, it is necessary to perform pre-processing to remove compression artifacts. We pre-process the training data in the same way as in [29], using only videos with both width and height of more than 720 pixels. Then, we downsample them to have 352×288 resolution for removal of compression artifacts. We train on 30K videos with 450K frames from the Kinetics video dataset. For evaluation, we test the BP-DVC Net on the raw video datasets such as Video Trace Library (VTL) [46],

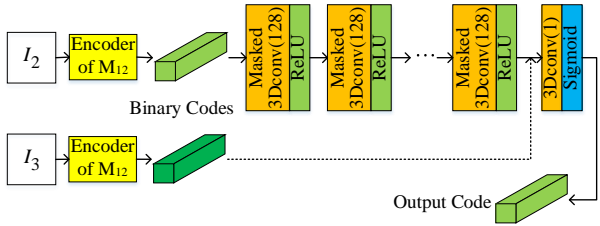

Figure 4. Proposed entropy coding model for the $M_{12}$ model.

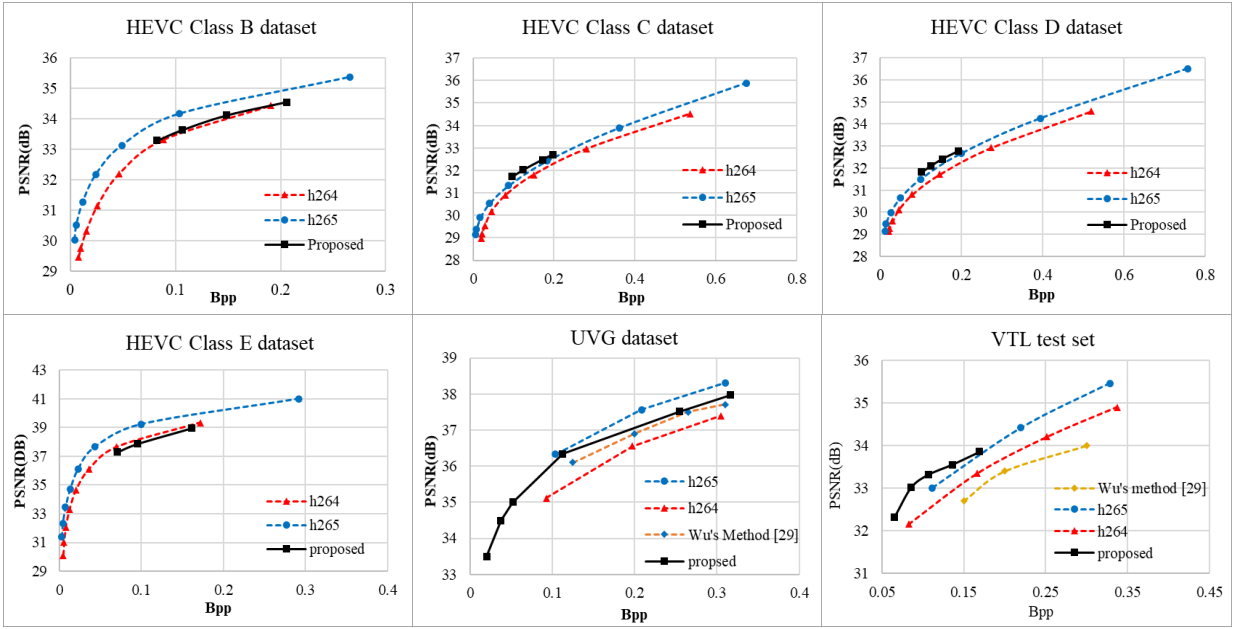

Figure 5. Comparison to the conventional video codecs such as H.264 and H.265, and the deep learning based video compression [29] in terms of PSNR. Our method outperforms H.264 and Wu's method [29] for most datasets in terms of PSNR. Also, our method has similar or better compression performance compared with H.265 in terms of PSNR.

Ultra Video Group (UVG) [47] and the HEVC Standard Test Sequences (Class B, C, D and E) [2]. The VTL dataset contains the videos with a size of 352×288. The UVG dataset contains the videos with a size of 1920×1080. The videos in the HEVC dataset has a different size depending on the class type.

**Implementation details.** The proposed BP-DVC Net is trained based on the total loss $l$ as:

$$l = m_a l_a + m_p l_p + m_f l_f \quad (9)$$

where $l_a$ is the auto-encoder loss, $l_p$ is the bidirectional prediction loss and $l_f$ is the flow loss with their coefficients ($m_a = 1$, $m_p = 1$, and $m_f = 0.1$ in our experiments). Since it is difficult to train all the networks of our BP-DVC Net at once, we propose a training strategy suitable for the proposed method. We first pre-train the optical flow estimation network by training the flow loss during 50K iterations. Then, we train the bidirectional prediction network with the prediction loss and the flow loss up to 250K iterations. Finally, the BP-DVC Net is trained up to 500K iterations with the total loss using ADAM [48] with the initial learning rate 0.0005, which is divided by 2 for every 100K iterations. All models ($M_{12}$, $M_{33}$, and $M_{66}$) are trained with 10 iterations of the LSTM network. Batch normalization is not used and gradient norm clipping is used with 0.5 for our model. For training, we used a batch size of 16 and a patch of 64×64 randomly cropped.

**Evaluation.** We measure both distortion and bitrate simultaneously. Two metrics such as PSNR and the Multi-Scale Structural Similarity Index (MS-SSIM) [49] are used to measure distortion. We use bits per pixel (Bpp) to measure the bitrates. The reconstruction with lower Bpp, higher PSNR and MS-SSIM is the better.

### 4.2. Experimental Results

The BP-DVC Net is compared with the conventional video codecs such as AVC/H.264 and HEVC, as well as the deep learning based video compression model in [29]. For fair comparison, the GOP size of the conventional video compression codecs is fixed to 12, and we use the same setting of the conventional video codecs as in [31]. Fig. 5 shows the rate-distortion (R-D) curves for the VTL, UVG, and HEVC datasets. Our BP-DVC Net outperforms H.264 and Wu's method [29] in terms of PSNR. More experimental results are provided in the supplemental material. Our method has a limitation at the moment that P-frame prediction coding is not incorporated, compared with the conventional video codecs.

### 4.3. Ablation Study

In the proposed method, the optical flow estimation network with the flow loss and the bidirectional prediction network with the prediction loss are the key components in achieving higher compression performance. In order to demonstrate the contribution of each component, we have the experiments of excluding components one by one from the entire structure as shown in Fig. 6.

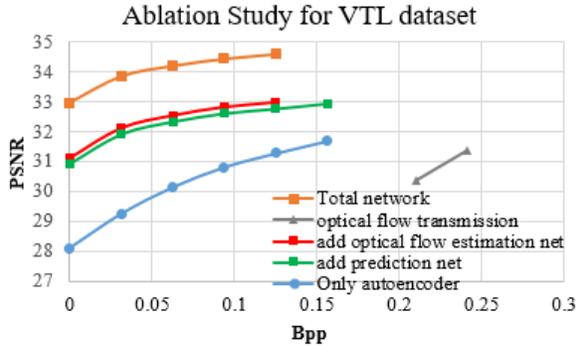

Figure 6. Ablation study for VTL dataset with $M_{33}$ model.

**Optical flow estimation network.** Optical flow estimation networks have significant coding efficiency improvements over optical transmission networks. This means that optical flow estimation works well without transmitting the motion information to the decoder sides.

**B-frame prediction network.** The bi-directional prediction network shows larger influences on compression performance than the auto-encoder alone. This shows that the bi-directional predication network focuses on the prediction for the current frame, allowing the auto-encoder to focus on residual compression.

**Proposed entropy coding model.** The existing entropy coding model used in [29] has a coding efficiency improvement of about 2% in our entire network. The proposed entropy coding model has a coding efficiency improvement of about 5% using the temporal context of the neighboring frames.

## 5. Conclusion

Our proposed BP-DVC Net is the deep end-to-end video compression model based on the bidirectional prediction network. In particular, we proposed a bidirectional prediction network and an optical flow estimation network, and improved compression performance of the base auto-encoder by introducing the prediction loss and flow loss. In addition, the temporal context of the neighboring frames' features is applied to the proposed entropy coding model to improve the compression performance slightly. The proposed video compression scheme showed better compression performance than the existing video compression codecs. Since our work currently uses the fixed I frame and B frame in combination, the compression performance may be limited. Therefore, the future work will find a better way to find the optimal combination of I-, P-, and B-frame combinations and apply it to a deep learning based network.

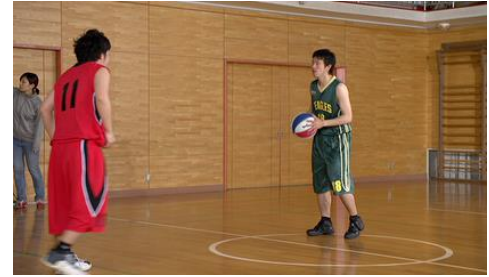

(a) 12$^{th}$ frame of BasketballPass

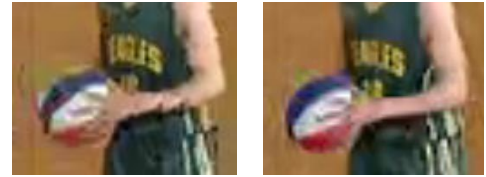

(b) H.264  (c) H.265

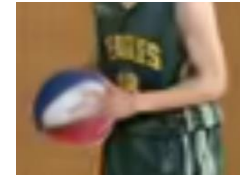

(d) Proposed method

Figure 7. The comparison of subjective quality results with H.264 and HEVC (H.265) at about 0.08 bpp.

References

[1] T. Wiegand, G. J. Sullivan, G. Bjontegaard, and A. Luthra, "Overview of the h.264/avc video coding standard," *IEEE Transactions on circuits and systems for video technology*, vol. 13, no. 7, pp. 560-576, July 2003.

[2] G. J. Sullivan, J.–R. Ohm, W.-J. Han, and T. Wiegand, "Overview of the high efficiency video coding (hevc) standard," *IEEE Transactions on circuits and systems for video technology*, vol. 22, no. 12, pp. 1649-1668, Dec. 2012.

[3] D. Mukherjee, J. Bankoski, A. Grange, J. Han, J. Koleszar, P. Wilkins, Y. Xu, and R. Bultje, "The latest open-source video codec VP9 – an overview and preliminary results," *2013 IEEE Picture Coding Symposium (PCS)*, pp. 390-393, Dec. 2013.

[4] C. Dong, C. C. Loy, K. He, and X. Tang, "Learning a deep convolutional network for image super-resolution," *European conference on computer vision*, Springer, pp. 184-199, Sep. 2014.

[5] J. Kim, J.K. Lee, and K.M. Lee, "Accurate image super-resolution using very deep convolutional networks," *Proceedings of the IEEE conference on Computer Vision and Pattern Recognition*, pp. 1646-1654, 2016.

[6] B. Lim, S. Son, H. Kim, S. Nah, and K.M. Lee, "Enhanced deep residual networks for single image super-resolution," *Proceedings of the IEEE conference on Computer Vision and Pattern Recognition Workshops*, pp. 136-144, 2017.

[7] C. Ledig, L. Thesis, F. Huszár, J. Caballero, A. Cunningham, A. Acosta, A. Aitken, A. Tejani, J. Totz, Z. Wang, and W.

Shi, "Photo-realistic single image super-resolution using a generative adversarial network," *Proceedings of the IEEE conference on Computer Vision and Pattern Recognition*, pp. 4681-4690, 2017.

[8] S. Niklaus, L. Mai, and F. Liu, "Video frame interpolation via adaptive convolution," *Proceedings of the IEEE conference on Computer Vision and Pattern Recognition*, pp. 670-679, 2017.

[9] S. Niklaus, L. Mai, and F. Liu, "Video frame interpolation via adaptive separable convolution," *Proceedings of the IEEE International Conference on Computer Vision*, pp. 261-270, 2017.

[10] S. Meyer, A. Djelouah, B. McWilliams, A. Sorkine-Hornung, M. Gross, and C. Schroers, "PhaseNet for video frame interpolation," *Proceedings of the IEEE conference on Computer Vision and Pattern Recognition*, pp. 498-507, 2018.

[11] S. Niklaus and F. Liu, "Context-aware synthesis for video frame interpolation," *Proceedings of the IEEE conference on Computer Vision and Pattern Recognition*, pp. 1701-1710, 2018.

[12] H. Jiang, D. Sun, V. Jampani, M.-H. Yang, E. Learned-Miller, and J. Kautz, "Super SloMo: high quality estimation of multiple intermediate frames for video interpolation," *Proceedings of the IEEE conference on Computer Vision and Pattern Recognition*, pp. 9000-9008, 2018.

[13] N. Johnston, D. Vincent, D. Minnen, M. Covel, S. Singh, T. Chinen, S. J. Hwang, J. Shor, and G. Toderici, "Improved lossy image compression with priming and spatially adaptive bit rates for recurrent networks," *Proceedings of the IEEE conference on Computer Vision and Pattern Recognition*, pp. 4385-4393, 2018.

[14] G. Toderici, S. M. O'Malley, S. J. Hwang, D. Vincent, D. Minnen, S. Baluja, M. Covell, and R. Sukthankar, "Variable rate image compression with recurrent neural networks," *4th International Conference on Learning Representation*, 2016.

[15] G. Toderici, D. Vincent, N. Johnston, S. J. Hwang, D. Minnen, J. Shor, and M. Covell, "Full resolution image compression with recurrent neural networks," *Proceedings of the IEEE conference on Computer Vision and Pattern Recognition*, pp. 5306-5314, 2017.

[16] J. Ballé, V. Laparra, E. P. Simoncelli, "End-to-end optimized image compression," *5th International Conference on Learning Representations*, 2017.

[17] J. Ballé, D. Minnen, S. Singh, S. J. Mwang, and N. Johnston, "Variational image compression with a scale hyperprior," *6th International Conference on Learning Representations*, 2018.

[18] M. Li, W. Zuo, S. Gu, D. Zhao, and D. Zhang, "Learning convolutional networks for content-weighted image compression," *Proceedings of the IEEE conference on Computer Vision and Pattern Recognition*, pp. 3214-3223, 2018.

[19] J. Lee, S. Cho, S.-K. Beack, "Context-adaptive entropy model for end-to-end optimized image compression," *arXiv preprint arXiv:1809.10452*, 2018.

[20] F. Mentzer, E. Agustsson, M. Tschannen, R. Timofte, and L. V. Gool, "Conditional probability models for deep image compression," *Proceedings of the IEEE conference on Computer Vision and Pattern Recognition*, pp. 4394-4402, 2018.

[21] O. Rippel and L. Bourdev, "Real-time adaptive image compression," *Proceedings of the 34th International Conference on Machine Learning*, vol. 70, pp. 2922-2930, 2017.

[22] L. Thesis, W. Shi, A. Cunningham, and F. Huszár, "Lossy image compression with compressive autoencoders," *5th International Conference on Learning Representations*, 2017.

[23] G. K. Wallace, "The JPEG still picture compression standard," *IEEE trans. on consumer electronic*, vol. 38, no. 1, pp. xviii-xxxiv, Feb. 1992.

[24] A. Skodras, C. Christopoulos, and T. Ebrahimi, "The JPEG 2000 still image compression standard," *IEEE Signal Processing Magazine*, vol. 18, no. 5, pp. 36-58, Sep. 2001.

[25] F. Bellard, "Bpg image format," 2014, URL: http://bellard.org/bpg/.

[26] W. S. Park and M. Kim, "CNN-based in-loop filtering for coding efficiency improvement," *2016 IEEE 12th Image, Video, and Multidimensional Signal Processing Workshop (IVMSP)*, pp. 1-5, 2016.

[27] W. Cui, T. Zhang, S. Zhang, F. Jiang, W. Zuo, Z. Wan, and D. Zhao, "Convolutional neural networks based intra prediction for HEVC," *2017 Data Compression Conference*, Apr. 2017.

[28] Z. Zhao, S. Wang, S. Wang, X. Zhang, S. Ma, and J. Yang, "Enhanced bi-prediction with convolutional neural network for high efficiency video coding," *IEEE Trans. on Circuits and Systems for Video Technology*, Oct. 2018.

[29] C.-Y. Wu, N. Singhal, and P. Krähenbühl, "Video compression through image interpolation," *Proceedings of the European Conference on Computer Vision (ECCV)*, pp. 416-431, 2018.

[30] O. Rippel, S. Nair, C. Le, S. Branson, A. G. Anderson, and L. Bourdev, "Learned video compression," *arXiv preprint arXiv:1811.06981*, 2018.

[31] G. Lu, W. Ouyang, D. Xu, X. Zhang, C. Cai, and Z. Gao, "DVC: an end-to-end deep video compression framework," *arXiv preprint arXiv:1812.00101*, 2018.

[32] C. Jia, S. Wang, X. Zhang, S. Wang, J. Liu, S. Pu, and S. Ma, "Content-aware convolutional neural network for in-loop filtering in high efficiency video coding," *IEEE Trans. on Image Processing*, 2019.

[33] J. Kang, S. Kim, and K. M. Lee "Multi-modal/multi-scale convolutional neural network based in-loop filter design for next generation video codec," *IEEE International Conference on Image Processing*, Sep. 2017.

[34] Y. Zhang, T. Shen, X. Ji, Y. Zhang, R. Xiong, and Q. Dai "Residual Highway Convolutional Neural Networks for in-loop Filtering in HEVC," *IEEE Trans. on Image Processing*, Vol. 27, No. 8, Aug. 2018.

[35] C. Li, L. Song, R. Xie, and W. Zhang "CNN based post-processing to improve HEVC," *IEEE International Conference on Image Processing*, Sep. 2017.

[36] R. Yang, M. Xu, Z. Wang, and T. Li "Multi-Frame Quality Enhancement for Compressed Video," *Proceedings of the IEEE conference on Computer Vision and Pattern Recognition*, pp. 6664-6673, 2018.

[37] A. Dosovitskiy, P. Fischery, E. Ilg, C. Hazirbas, V. Golkov, P. van der Smagt, D. Cremers, and T. Brox, "FlowNet: Learning optical flow with convolutional networks," *IEEE International Conference on Computer Vision*, pp. 2758-2766, 2015.

[38] O. Ronneberger, P. Fischer, and T. Brox, "U-Net: Convolutional networks for biomedical image segmentation," *International Conference on Medical Image Computing and Computer Assisted Intervention (MICCAI)*, pp. 234-241, 2015.

[39] A. Ranjan and M. J. Black, "Optical flow estimation using a spatial pyramid network," *IEEE Conference on Computer Vision and Pattern Recognition*, pp. 4161-4170, 2017.

[40] D. Sun, X. Yang, M.-Y. Liu, J. Kautz, "PWC-Net: CNNs for optical flow using pyramid, warping, and cost volume," *IEEE Conference on Computer Vision and Pattern Recognition*, pp. 8934-8943, 2018.

[41] K. He, X. Zhang, S. Ren, and J. Sun, "Deep residual learning for image recognition," *IEEE Conference on Computer Vision and Pattern Recognition*, pp. 770-778, 2016.

[42] T. Zhou, S. Tulsiani, W. Sun, J. Malik, and A. A. Efros, "View synthesis by appearance flow," *European Conference on Computer Vision*, pp. 286-301, 2016.

[43] G. Farnebäck, "Two-frame motion estimation based on polynomial expansion," *Scandinavian conference on Image analysis*, pp. 363-370, 2003.

[44] WebP. https://developers.google.com/speed/webp/, accessed: 2019-03-21

[45] J. Carreira and A. Zisserman, "Quos vadis, action recognition? A new model and the kinetics dataset," *IEEE Conference on Computer Vision and Pattern Recognition*, pp. 6299-6308, 2017.

[46] Video trace library. http://trace.eas.asu.edu/index.html, accessed: 2019-03-21

[47] Ultra video group test sequences. http://ultravideo.cs.tut.fi, accessed: 2019-03-21

[48] D. P. Kingma and J. Ba, "Adam: a method for stochastic optimization," *arXiv preprint arXiv:1412.6980*, 2014.

[49] Z. Wang, E. P. Simoncelli, and A. C. Bovik, "Multiscale structural similarity for image quality assessment," *The Thirty-Seventh Asilomar Conf. on Signals, Systems & Computers*, vol. 2, pp. 1398-1402, 2003.

# Appendix

## 1. The proposed optical flow refinement sub-net

In the main paper, we described our proposed optical flow estimation network architecture. In order to avoid sending the motion information to the decoder sides, we estimates the optical flows between the current frame ($I_n$) and two decoded neighboring frames ($\hat{I}_{n-n_1}$ and $\hat{I}_{n+n_2}$) indirectly using $F_{(n-n_1)\to(n+n_2)}$ and $F_{(n+n_2)\to(n-n_1)}$. However, the approximated optical flows $\breve{F}_{(n)\to(n-n_1)}$ and $\breve{F}_{(n)\to(n+n_2)}$ may be less accurate when the motion is large. Therefore, the approximated optical flows are refined through the proposed optical flow refinement sub-net to improve the accuracy as depicted in Fig. A1. Fig. A1 shows our proposed optical flow refinement sub-net based on a U-Net structure [38]. We proposed the optical flow refinement sub-net that we utilize not only two skip connections between interim convolution layers in a U-Net structure but also the other skip connection between input and output of this network for predicting only the residual of the optical flows.

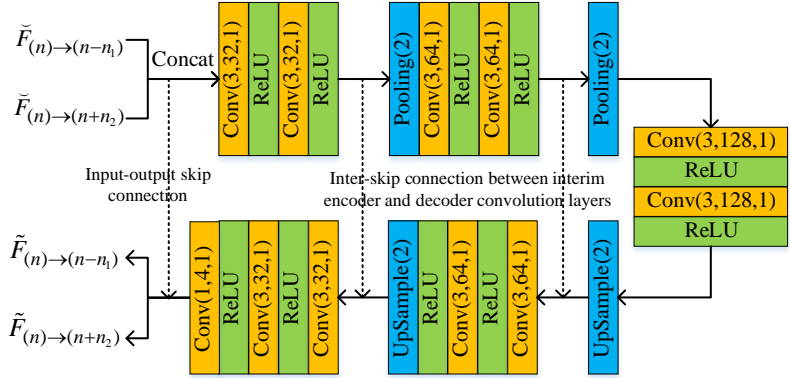

Figure A1. The proposed optical flow refinement sub-net based on a U-Net structure [38].

## 2. The proposed bi-directional prediction network

In the main paper, we described the overall structure of our proposed BP-DVC Net that consists of the optical flow estimation network, the bi-directional prediction network and ConvLSTM-based auto-encoder network. In particular, Fig. A2 shows the details of the proposed bi-directional prediction network based on a U-Net structure [38].

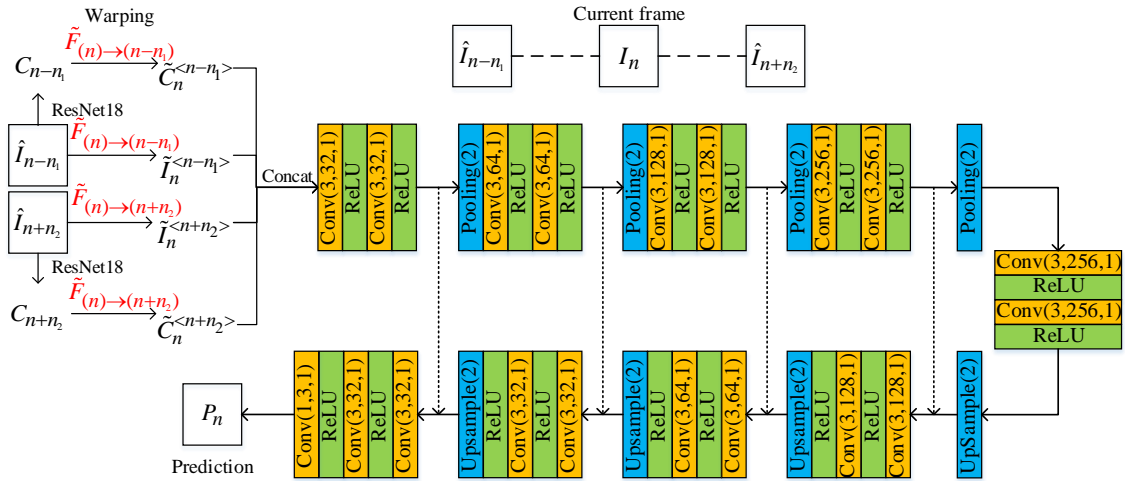

Figure. A2. The proposed bi-directional prediction network based on a U-Net structure [38].

The proposed bi-directional network predicts the current frame $I_n$ using two decoded neighboring frames, $\hat{I}_{n-n_1}$ and $\hat{I}_{n+n_2}$. For this, we first warp the two neighboring frames using their respective optical flows $\tilde{F}_{(n)\to(n-n_1)}$ and $\tilde{F}_{(n)\to(n+n_2)}$ obtained from the optical flow estimation network. In addition, the context information obtained from the first feature map of ResNet18 [41] for the two decoded neighboring frames are also warped using the same optical flows. Finally, we concatenate both of the warped two neighboring frames and the warped context feature maps and feed them into the bi-directional prediction network. Then, the bi-directional prediction network is trained to yield the output close to the current frame $I_n$.

## 3. The performance comparisons of R-D curves in terms of MS-SSIM

In the main paper, we showed the results of the R-D curves for our BP-DVC Net, AVC/H.264, HEVC, and Wu's method in terms of PSNR for the VTL [46], UVG [47], and HEVC [2] datasets. Fig. A3 shows the results of the R-D curves in terms of MS-SSIM for the same datasets as in the main paper. Our BP-DVC Net outperforms H.264, H.265 and Wu's method [29] in terms of MS-SSIM although we trained the BP-DVC Net only using L1 loss. Furthermore, the BP-DVC Net even outperformed HEVC in terms of MS-SSIM-versus-rate curves for the datasets for which it showed inferior results in PSNR-versus-rate curves. It can be noted from the results our BP-DVC Net can yield comparable or better subjective qualities of decoded images compared to HEVC.

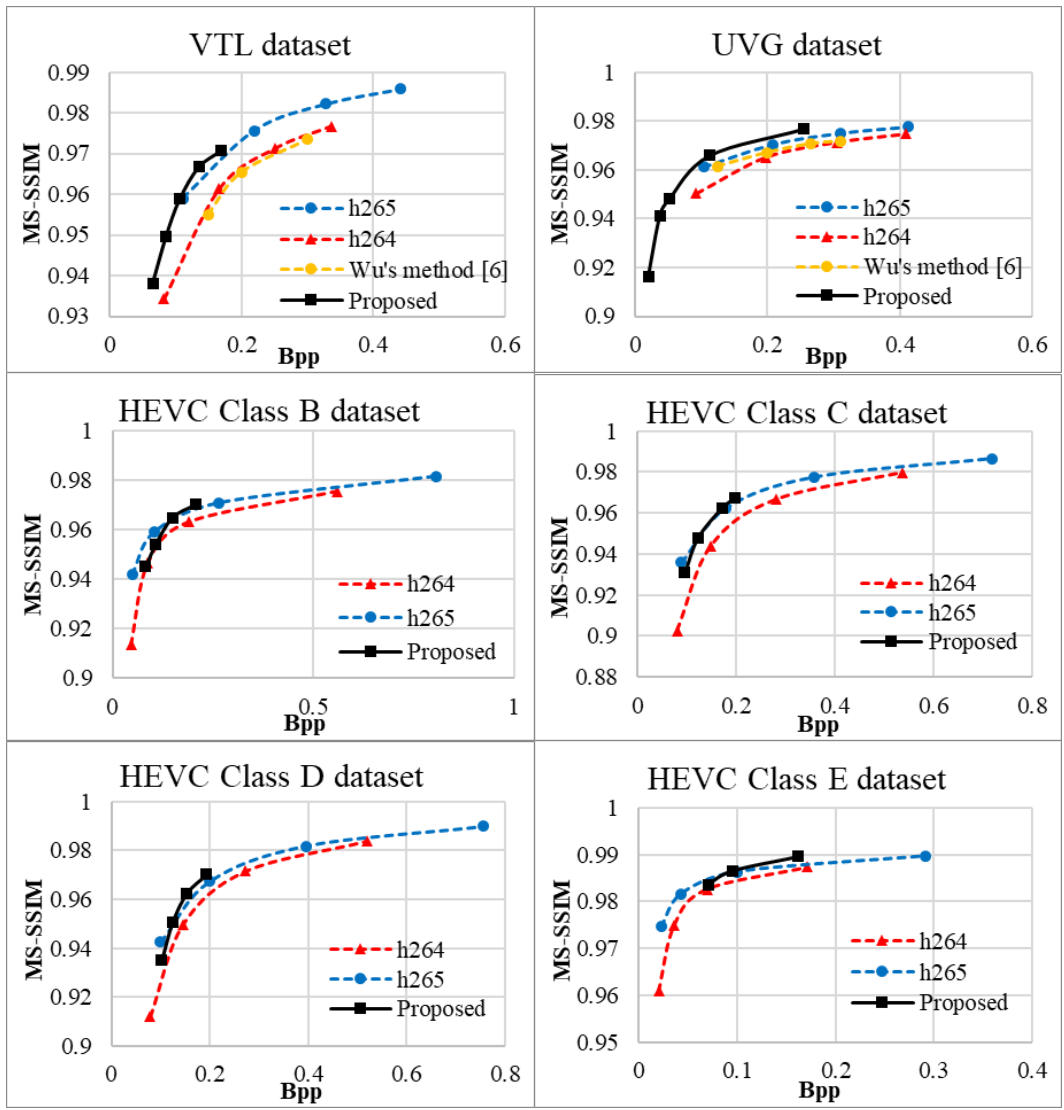

Figure A3. Comparison to the conventional video codecs such as H.264 [1] and H.265 [2], and the deep learning based video compression [29] in terms of MS-SSIM.

## 4. The comparison of subjective quality with both H264 [1] and H265 [2]

In the main paper, we provided the subjective quality comparison for our BP-DVC Net against the AVC/H264 and HEVC for *BasketballPass* sequence. In addition, we provide the subjective image quality comparisons for *Foreman* and *YachtRide* sequences, which are shown in Fig. A4 and A5, respectively. As can be observed in Fig. A4 and Fig. A5, the reconstructed frames by our BP-DVC Net show less coding artifact without blocking artifacts while the AVC/H264 exhibit severe both ringing and blocking artifacts and HEVC shows severe ringing artifacts. On the other hand, as shown in Fig. A4, the BP-DVC Net shows remarkably sharp edges around the auricle and helmet's brim while AVC/H.264 and HEVC show severe distortions. Similar results can also be observed in Fig. A5.

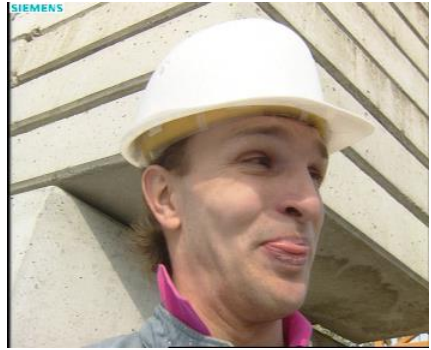

(a) Original frame of *Foreman* from VTL dataset

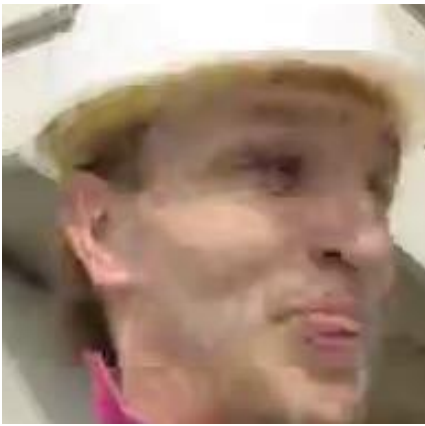 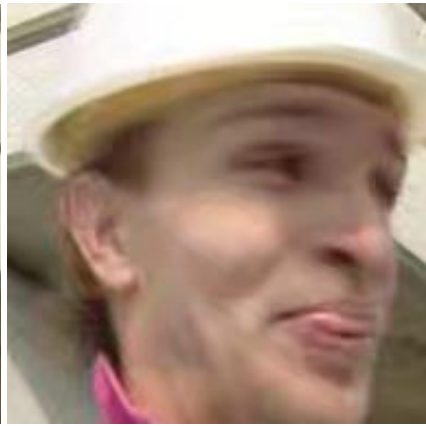 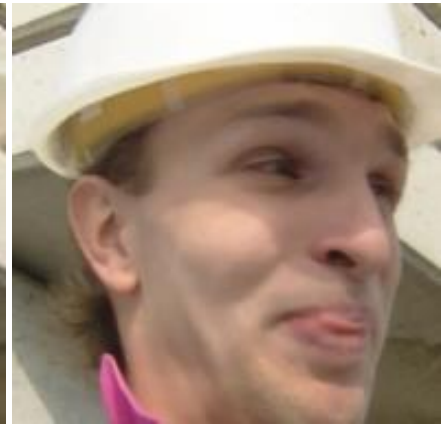

(b) H.264 (0.081 bpp)　　　　　(c) H.265 (0.086 bpp)　　　　　(d) Proposed (0.078 bpp)

Figure A4. The comparison of subjective quality with H.264 and H.265 for a *Foreman* sequence of VTL [4] dataset

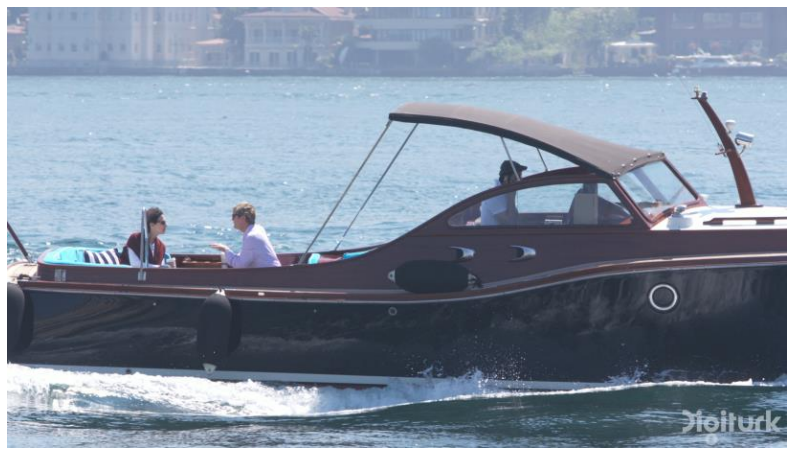

(a) Original frame of *YachtRide* from UVG dataset

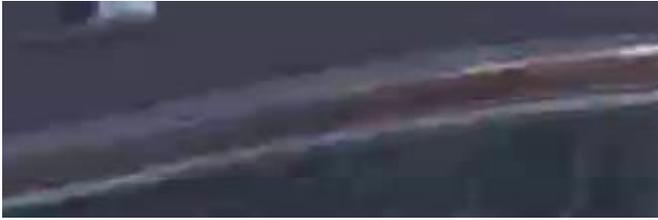

(b) H.264 (0.049 bpp)

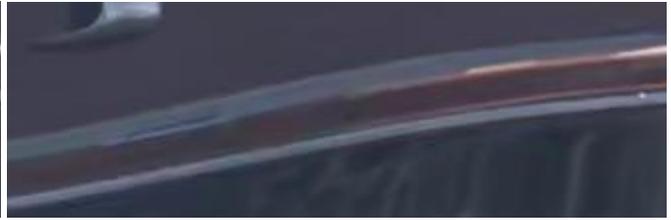

(c) H.265 (0.054 bpp)

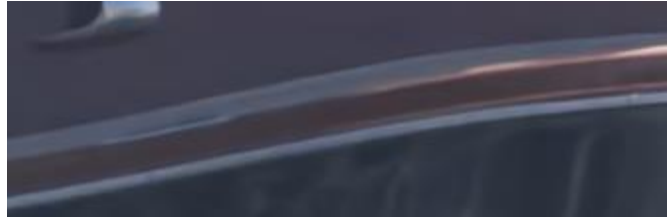

(d) Proposed (0.052 bpp)

Figure A5. The comparison of subjective quality with H.264 and H.265 for a *YachtRide* from UVG [5] dataset.